# FOUNDATION MODELS IN DERMATOPATHOLOGY: SKIN TISSUE CLASSIFICATION

Riya Gupta[1*†], Yiwei Zong[1†], Dennis H. Murphree[2*]
[1]Harvard T.H. Chan School of Public Health, Boston, MA
[2]Mayo Clinic, Rochester, MN, USA
*Co-corresponding authors: [priyalgupta@hsph.harvard.edu], [Murphree.Dennis@mayo.edu]
†Co-first authors

## ABSTRACT

The rapid generation of whole-slide images (WSIs) in dermatopathology necessitates automated methods for efficient processing and accurate classification. This study evaluates the performance of two foundation models, UNI and Virchow2, as feature extractors for classifying WSIs into three diagnostic categories: melanocytic, basaloid, and squamous lesions. [1; 2] Patch-level embeddings were aggregated into slide-level features using a mean-aggregation strategy and subsequently used to train multiple machine learning classifiers, including logistic regression, gradient-boosted trees, and random forest models. Performance was assessed using precision, recall, true positive rate, false positive rate, and the area under the receiver operating characteristic curve (AUROC) on the test set.

Results demonstrate that patch-level features extracted using Virchow2 outperformed those extracted via UNI across most slide-level classifiers, with logistic regression achieving the highest accuracy (90%) for Virchow2, though the difference was not statistically significant. The study also explored data augmentation techniques and image normalization to enhance model robustness and generalizability. The mean-aggregation approach provided reliable slide-level feature representations. All experimental results and metrics were tracked and visualized using WandB.ai, facilitating reproducibility and interpretability.[3]

This research highlights the potential of foundation models for automated WSI classification, providing a scalable and effective approach for dermatopathological diagnosis while paving the way for future advancements in slide-level representation learning.

# INTRODUCTION

In recent years, the generation of whole-slide images (WSIs) has increased significantly, creating a demand for automated WSI processing facilities to streamline file sorting and diagnostics. Previous work has demonstrated the efficacy of deep learning models in medical image analysis, with convolutional neural networks and transformers achieving high success in classification tasks. However, limited work has focused on leveraging foundation models specifically for dermatopathology. This project aims to develop a deep learning based approach for classifying WSIs of skin tissues into three distinct categories: melanocytic, basaloid, and squamous lesions. Such classification is essential for improving diagnostic efficiency and accuracy in dermatopathology.

We aim to assess the applicability of two foundational models, UNI and Virchow 2, within the domain of dermatopathology.[1; 2] Specifically, the project seeks to compare and contrast these existing models to determine which achieves superior performance on slides related to skin disease. This work represents the foundational step in a broader pipeline to develop targeted models for various dermatopathology-related tasks.

These downstream applications will encompass diagnosis, prognosis, treatment planning, practice efficiency, and quality control. By selecting the best foundation model, this project will establish a critical groundwork for building robust tools that enhance clinical workflows and support pathologists in their diagnostic processes.

# BACKGROUND

Foundation models are large pre-trained neural networks designed to extract meaningful features from data and can be fine-tuned for specific tasks. In digital pathology, these models offer a promising solution for automating the analysis of WSIs. This study focuses on two foundation models, UNI and Virchow2, both designed for medical imaging tasks and use Vision Transformer (ViT) architectures to process images as patches.[1; 2; 4] After an image is segmented into patches, ViT processes each patch as a vector by flattening its pixel values and embedding it with positional information. These patch embeddings are then passed through transformer layers to learn both local and global relationships within the image.[4]

UNI is based on the ViT-L (Large) architecture and trained on the Mass-100K dataset, which includes 100,000 WSIs from Brigham and Women's Hospital and Massachusetts General Hospital.[1] Virchow2 uses the ViT-H (Huge) architecture and was trained on a larger dataset of 3.1 million WSIs from Memorial Sloan Kettering Cancer Center.[2] Both models use self-supervised learning via DINOv2 (Distillation with No Labels), which leverages masked image modeling to learn patterns in unlabeled data.[5] In this approach, portions of the input data are hidden (masked), and the model learns to predict the missing parts, enabling it to generate embeddings that capture essential features.

These models were selected due to their strong performance in extracting slide-level features and their ability to generalize to diverse medical imaging tasks. By comparing UNI and Virchow2 for classifying WSIs into three dermatopathological categories (melanocytic, basaloid, and squamous lesions), we aim to identify the most effective model for this application.

# METHODS

In this section, we will give a detailed description of the methods we have adopted to generate the final prediction of the skin concern of images at both patch level and slide level, using foundation models of interest – UNI and Virchow 2 – within the domain of dermatopathology. Specifically, we are interested in comparing and contrasting existing foundation models to find the best performance for slides related to skin disease. For this purpose, the first task is to preprocess the data in the desired format of direct input for the two foundation models. Then, we will introduce how the model is deployed on a secure institutional computing platform and how patch-level features are extracted using different models. Finally, the model evaluation and comparison routine are built based on functionalities in wandb.ai.[3]

## 2.1 *Data Acquisition*

The source data acquisition was conducted by Mayo's research staff. This process involved collecting various skin tissue slides and extracting patches from them. Initially, we obtained 960 WSIs for analysis. Among these WSIs, 126 slides were labeled as "basaloid," 263 slides as "melanocytic," 325 slides as "squamous," and 246 slides as "other". Since our research focuses only on the three categories, slides labeled as "other" were excluded from the analysis. The WSIs were labeled by board-certified dermatopathologists to ensure labeling accuracy.

These slides were then segmented into smaller patches, each measuring 512x512 pixels at a 40x magnification. This patch size and magnification were selected to capture sufficient histopathological detail while optimizing computational efficiency for downstream analysis. After segmentation, we collected the patches from the 960 tar files. Most slides generated a moderate number of patches, typically ranging from 1000 to 4000 patches per slide (for some particularly large-sized slides, the number of patches exceeds 20,000), resulting in a total of 3,348,152 patches.

**2.2 *Data description***

The WSI dataset is accompanied by a descriptive table summarizing the metadata for each WSI file. The table contains 960 rows and ten columns, each representing a unique slide and its associated metadata, see **Table 1** as an example. The variables in the table are defined below:
- **fullpath**: Full path to the zipped path file in the cloud storage.
- **file**: File name for the slide file.
- **case_id**: Unique identifier for each case.
- **sub_block and block**: Identification for subsections within cases.
- **DIAG_SCORE**: Diagnostic score for the slide file.
- **DIAG_TYPE**: Diagnostic type indicating the sub-classification label for the slide, providing a more nuanced classification than the primary categories.
- **category**: Type of skin concern (the three main categories: basaloid, squamous, and melanocytic, along with one additional category labeled as "other").
- **subset**: Indicates whether the slide belongs to the training, validation, or testing set.
- **Staining**: Staining protocol applied, such as Hematoxylin and Eosin.

For the slide classification task, the most critical metadata fields are the file name and its associated category, which provide the ground truth labels for each slide. These ground truth labels serve as the basis for evaluating the extracted image embeddings and the model's classification performance.

The metadata also includes the subset variable, which indicates the predetermined data partition (training, validation, or testing). The dataset was initially split into three subsets: 498 files (69.6%) for training, 108 files (15.2%) for validation, and 108 files (15.2%) for testing. However, as no standalone validation process was conducted for this task, the validation set was merged with the test set to streamline evaluation. This resulted in a final split of 498 files (69.6%) for training and 216 files (30.4%) for testing. This approach ensured a substantial portion of the data was used for training while preserving a meaningful portion for robust evaluation of the model's performance.

**Table 1**: Descriptive table summarizing metadata examples.

| fullpath | file | case_id | sub_block | block | DIAG_SCORE | DIAG_TYPE | category | subset | Staining |
|---|---|---|---|---|---|---|---|---|---|
| Masked[1] | Masked | Masked | B1-1 | B | 100 | squamous cell carcinoma in situ | squamous | Train | H&E |
| Masked | Masked | Masked | B1-2 | B | 100 | dermatitis | other | Train | H&E |
| Masked | Masked | Masked | E1-1 | E | 100 | junctional nevus with severe atypia | melanocytic | Train | H&E |
| Masked | Masked | Masked | D1-1 | D | 100 | basal cell carcinoma | basaloid | Train | H&E |
| Masked | Masked | Masked | B1-2 | B | 100 | basal cell carcinoma | basaloid | Test | H&E |

**2.3 *Data Missingness and Potential Limitations***

Upon verification, we found that only 847 tar files contained the patches, with 113 files being empty. This discrepancy suggests a potential issue during the slide segmentation process. To address this, we filtered out all the empty files to minimize any potential negative impact on model performance. Subsequently, the issue was resolved, and the 113 empty files were updated to include the missing patches.

Additionally, we considered potential limitations in the dataset, particularly the variability in staining protocols across slides, which may influence model training and affect the reproducibility of prediction results. All the slides in the dataset were stained using Hematoxylin and Eosin (H&E) from Mayo's own lab. However, we observed slight variations in staining intensity and appearance. These variations include standard H&E stains, H&E recuts (repeat stains on the same tissue), and H&E C-G (H&E stains with cytoplasmic granularity enhancements). Although the basic staining protocol remains consistent, it is important to note that these subtle differences in staining may impact the models' training and results. This variability is an inherent challenge in WSI datasets and will be accounted for in the analysis and interpretation of results.

**2.4 *Data Preprocessing***

During the data preprocessing phase, the tar files were copied to a designated directory, unzipped, and patches were extracted from the WSIs. The extracted patches were then stratified into three distinct datasets: training, validation, and testing subsets. This partitioning was essential for facilitating model development and evaluation. Each patch was subsequently labeled with the slide-level label into one of three categories: melanocytic, basaloid, or squamous. To improve the robustness and generalizability of the models, a series of data augmentation techniques and normalization steps were applied.

---

[1] *Masked data/information indicates that it is not publicly available due to proprietary or sensitive content.*

To ensure consistent input dimensions across the dataset, all extracted patches were resized to a uniform size of 224x224 pixels. This resizing process was implemented using the *torchvision.transforms* module, which provides an efficient and straightforward way to handle image transformations in PyTorch. The resizing preserved the aspect ratios through appropriate interpolation methods.

To introduce variability and improve model robustness, several image data augmentation techniques were employed, including horizontal and vertical flips, Gaussian noise, motion blur, median blur, normal blur, random affine transformations (translation, scaling, and rotation), and random adjustments to brightness and contrast. These augmentations generated variations in the dataset, making the model more adaptable to different input conditions.

Data normalization was performed on the patches to standardize their pixel value distributions. The normalization parameters matched the pre-training settings of the UNI model. Specifically, all patches were resized to a uniform size of 224x224 pixels with three RGB channels. The patches were then normalized to X` using the common mean values (0.485, 0.456, 0.406) and standard deviations (0.229, 0.224, 0.225), consistent with the UNI pre-training settings. This normalization was performed according to the formula:

$$X` = (X - \mu)/\sigma \tag{1}$$

where X represents the pixel values of the 224x224 images, μ denotes the mean values, and σ represents the standard deviations for the RGB channels. Each patch was subsequently converted into tensor format to ensure compatibility with PyTorch.

### 2.5 *Model Development*

In this project, both foundation models functioned as feature extractors. The pre-processed patch-level images (saved as 224x224 tensors) were fed into the foundation models, which generated equally sized features (also referred to as embeddings) for each patch.

A critical step in the learning process was accessing and acquiring the pre-trained weights for the foundation models. As both models are publicly available as open-source projects, this involved requesting access to the pre-trained weights from the respective model developers. Once the pre-trained weights were acquired, a Python-based functionality was implemented to extract embeddings for all patches segmented from a single WSI. These embeddings were stored in individual PyTorch tensors, along with the corresponding ground truth class labels and dataset subsets (training, validation, or testing). This information was cached in binary files for efficient storage and retrieval.

This workflow ensures that the model inference results are stored in a clear and consistent manner, facilitating subsequent slide-level classification tasks.

## 2.6 *Slide-level Classification*

After extracting patch-level embeddings, each WSI slide was associated with feature vectors, with the number of vectors varying based on the patches extracted from each slide. The next step involved aggregating these patch-level features to obtain consistent-sized, fixed-dimension features at the slide level.

In the original Virchow paper, the researchers proposed an advanced transformer-based patch-level embedding aggregation architecture called AGATA, which uses cross-attention to learn patch weights based on their contribution to the whole slide label decision.[6] However, neither the source code for the AGATA architecture nor its pre-trained weights were made publicly available for direct deployment. Additionally, AGATA requires precise spatial details of patch locations on slide-level images, which are unavailable in our dataset. As a result, a simpler mean-aggregation approach was adopted. This method calculates the average of all patch embeddings, aggregating slide-level features from dimensions [$m$, 1024] or [$m$, 1280] ($m$ is the number of patches per slide) to [1, 1024] for UNI or [1, 1280] for Virchow2, respectively. For the 714 WSIs (after excluding the 'other' class), the final design matrices were of dimensions [714, 1024] for UNI and [714, 1280] for Virchow2.[1; 2]

Using these design matrices, a range of common machine-learning models was applied to perform the final slide-level classification. The models included logistic regression, AdaBoost, Decision Tree, Gradient Boosting Machine (GBM), Random Forest, k-Nearest Neighbor (kNN), and Naive Bayes. All models were trained on the training subset, and their performance was evaluated on the testing subset. Hyperparameter combinations for these models (where applicable) were optimized using 5-fold cross-validation.

## 2.7 *Model Evaluation*

A comprehensive performance assessment was conducted for the slide-level classification task to evaluate the practical performance of the models in real-world applications. After hyperparameter tuning and final model training, predictions for the test set were generated using the optimized models. The performance of the slide-level classification was assessed using the following metrics: accuracy, F1 score, precision, recall, true positive rate (TPR), false positive rate (FPR), and the Area Under Receiver Operator Curve (AUROC). Precision and recall were particularly useful in evaluating the trade-off balance between false positives and false negatives, while TPR and FPR provided insights into the model's sensitivity and error patterns.

To centralize the tracking and visualization of evaluation metrics, we utilized WandB.ai, a widely used platform for experiment tracking in AI development.[3] WandB enabled the logging of precision-recall curves, and AUROC plots, providing clear insights into category-specific performance and highlighting any potential imbalances across diagnostic types. We also plan to plot the learning curves to provide some insights into the performance characteristics across varying training set sizes by WandB. The platform facilitated reproducibility and allowed for easy comparison between the two foundation models and the slide-level classifiers.

## 2.8 *Schematic*

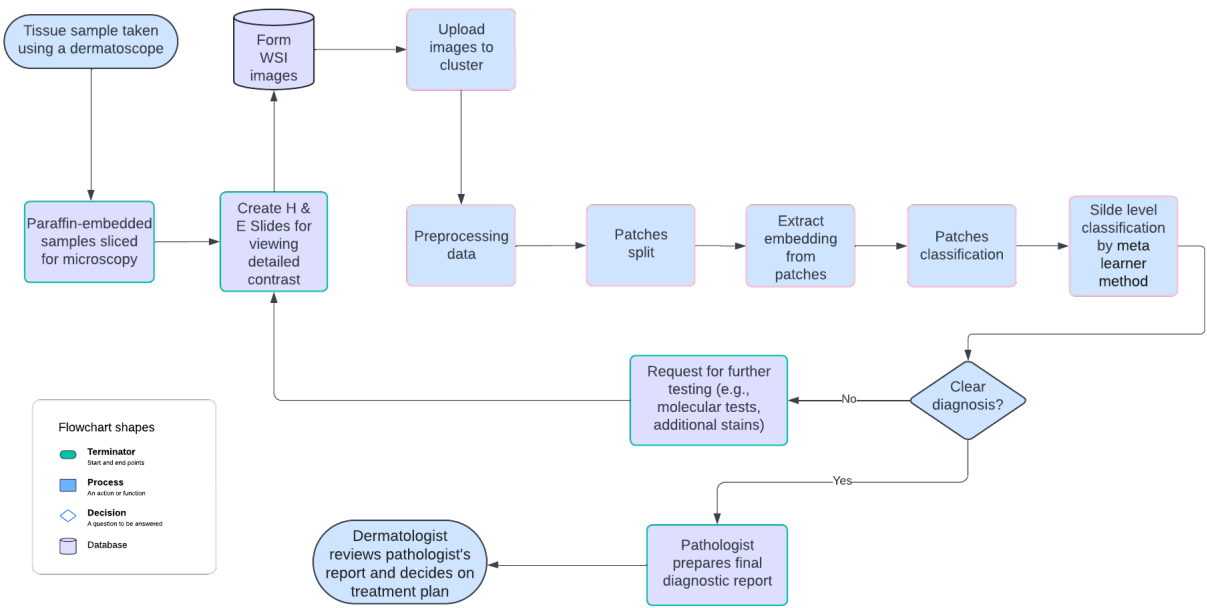

**Figure 1**: A schematic illustrating the data and information workflow

The schematic above outlines the data and information flow from input to output, highlighting the core components of this process. Initially, whole slide images are preprocessed and segmented into patches, which are then analyzed by the model to generate embeddings. These embeddings are classified into designated categories, providing slide-level classifications that are communicated back to the user interface. The model's outputs will be delivered to the user interface as a diagnostic report, which will include the category prediction (e.g., melanocytic, basal cell, or squamous), confidence scores (the confidence score for each classification), visualization images (a heatmap overlay highlighting areas relevant to the classification), and next-step recommendations.

Currently, the output report contains only the category predictions and an assessment of the associated results. The inclusion of confidence scores, visualization images, and next-step recommendations represent a vision for potential future work. This workflow ensures data integrity throughout, enabling accurate and efficient dermatopathology classification.

# RESULTS

### 3.1 *Exploratory Data Analysis*

This section presents the exploratory data analysis, highlighting key aspects of the WSI dataset. We remove the "other" category here as we only classify the three categories in this research.

The class distribution of the entire WSI dataset is visualized in **Figure 2**. The plot indicates that there is no significant class imbalance within the dataset. While "squamous" and "melanocytic" are more prevalent compared to "basaloid", all categories contain a sufficient number of samples to ensure adequate representation for model training and evaluation.

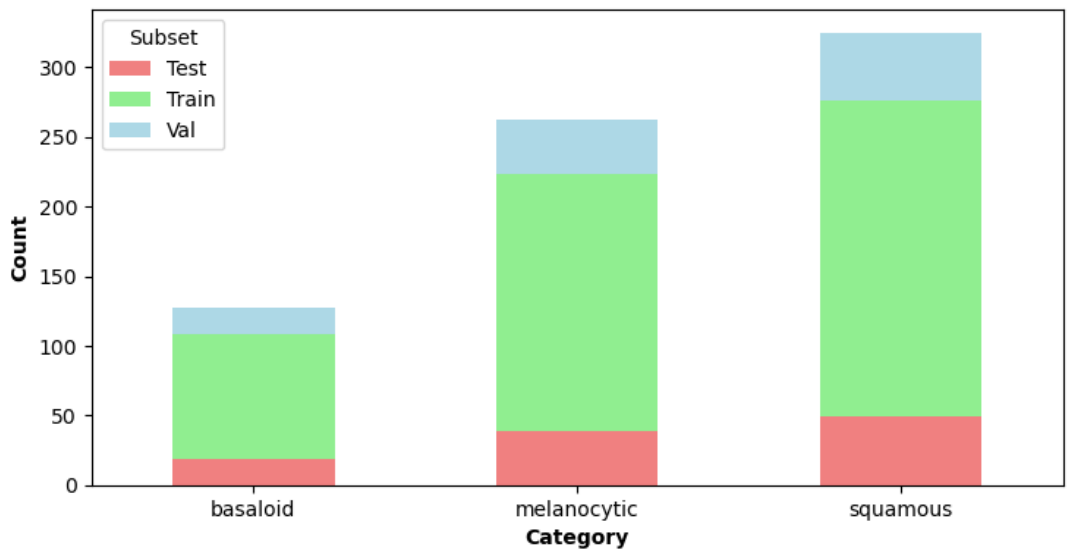

**Figure 2**: Class and subset composition of the dataset used in this project.

Following patch segmentation of the WSI files, **Figure 3** illustrates the distribution of the number of patches for a single WSI. The distribution is highly variable and right-skewed. Most WSIs generated between 1,000 to 5,000 patches, each measuring 512x512 pixels. However, a few outlier slides produced an exceptionally high number of patches (over 20,000 patches).

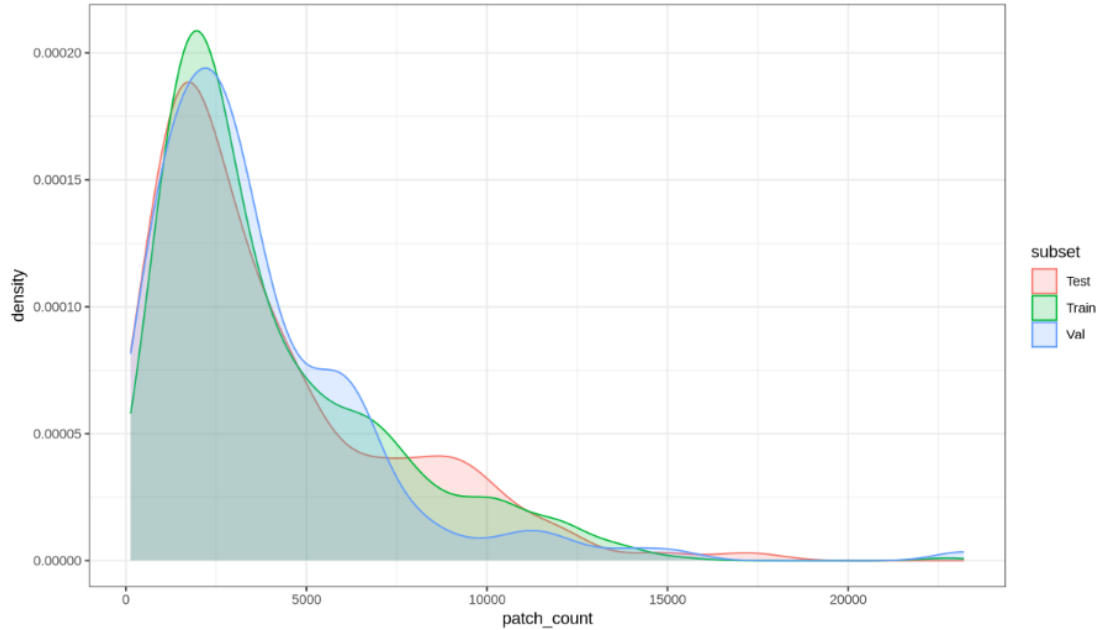

**Figure 3**: Distribution of the number of patches extracted for the WSIs.

To validate the patch segmentation process and gain an initial understanding of the extracted patches, a 3x5 grid was created, displaying three randomly selected patches from each of the classes of interest (**Figure 4**). All patches are of equal size and clearly show a focal region of the corresponding slide.

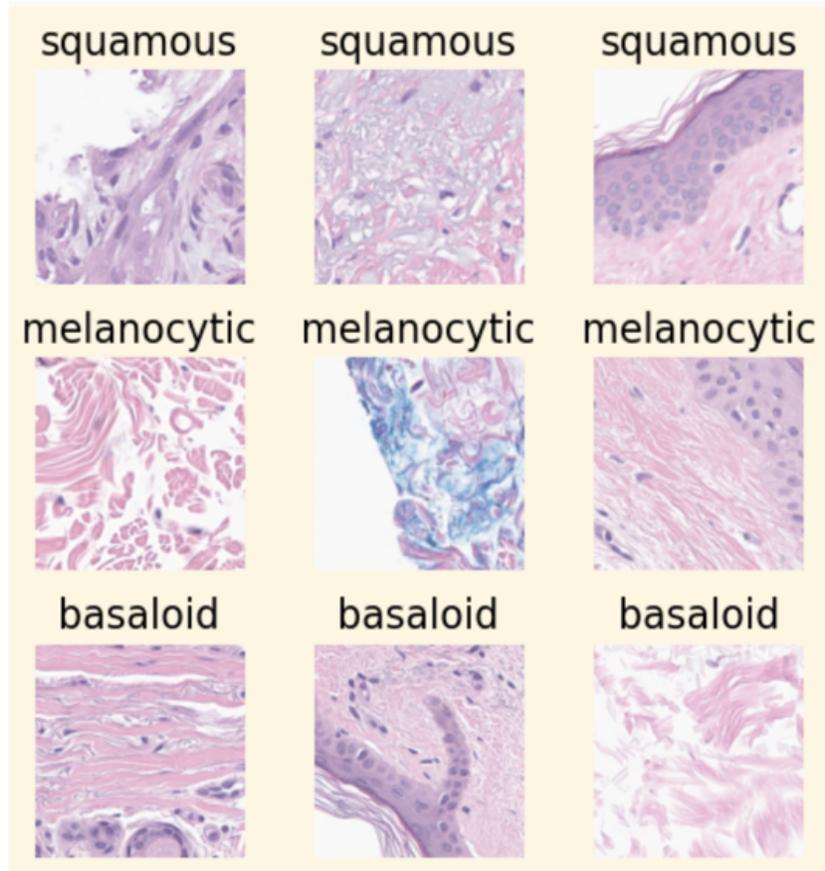

**Figure 4**: Three randomly selected patches from each class.

To explore the impact of data augmentation techniques on the patches, **Figure 5** presents a sample patch processed with various data augmentation transformations. These random transformations enhance the model's ability to generalize and improve robustness to variations in rotation, position, and resolution of the input images.

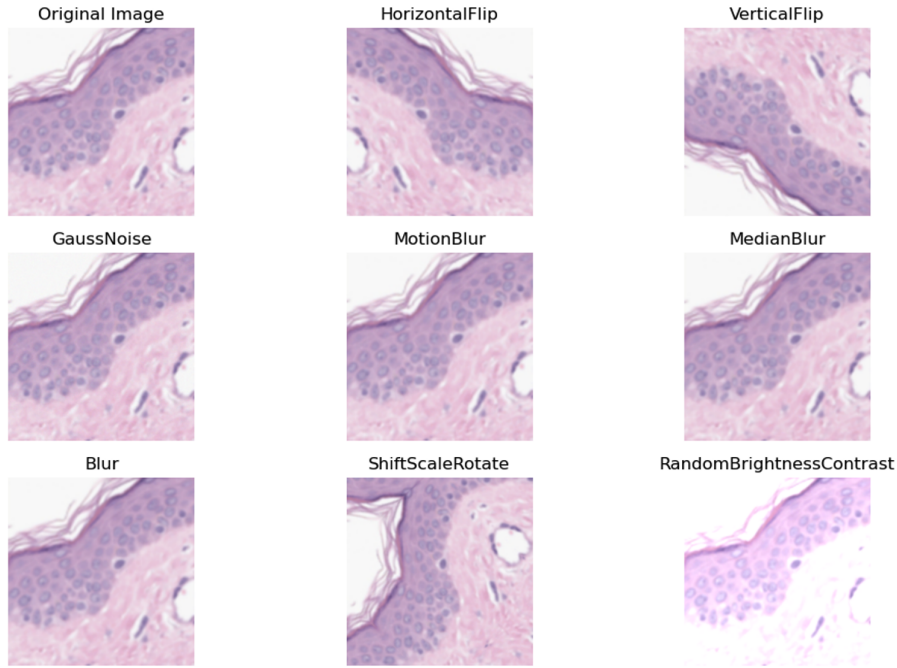

**Figure 5**: Effects of different data augmentation techniques on the same patches.

Finally, image pixel value normalization was performed using the weights described in the Methods section, and the resulting images were visualized in **Figure 6**, showing the same set of patches as in **Figure 4**. After normalization, the pixel values are no longer in their original scale, resulting in images without normal color mapping or proper contrast. However, the mean and standard deviations of the pixel values across the three channels were verified and found to be close to the theoretic values (mean close to 0, while standard deviation close to 1).

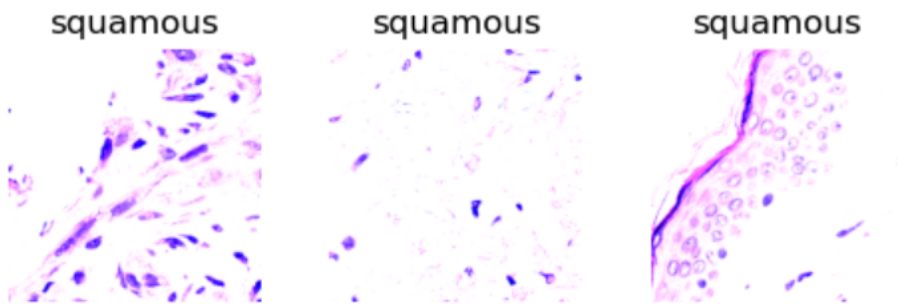

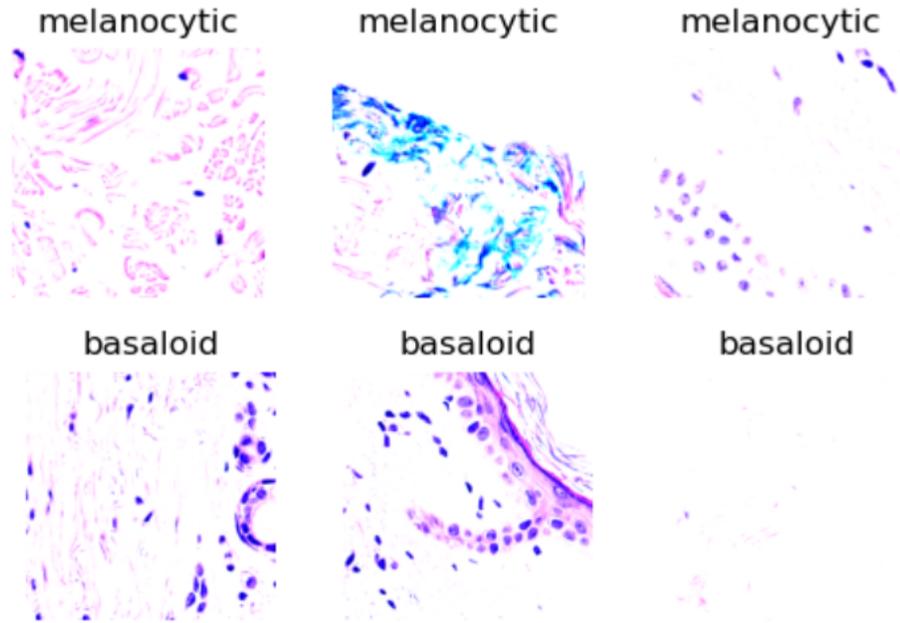

**Figure 6**: Normalized versions of the patches shown in **Figure 4**.

### 3.2 *Model Evaluation and Comparison*

After obtaining slide-level embeddings from both foundation models and training various machine learning models on top of them, this section presents a comprehensive comparison of model performances. The evaluation was conducted on the test set using the metrics outlined in the Methods section.

**Table 2**: Performance comparison of classifiers on slide-level embeddings from UNI and Virchow2 foundation models (accuracy in percentage).

| Classifier | Accuracy (UNI) | Accuracy (Virchow2) |
|---|---|---|
| k-Nearest Neighbor (kNN) | 83% | 84% |
| Decision Tree | 76% | 77% |
| Gradient Boosting | 87% | 89% |
| Random Forest | 84% | 88% |
| Logistic Regression | 88% | 90% |
| Naive Bayes | 80% | 86% |
| AdaBoost | 82% | 86% |

**Table 2** above shows the accuracy of classifiers across foundational models. While Virchow2 shows slight improvements over UNI across all seven classifiers, with differences ranging from 1% to 6%, it is important to note that these differences are not statistically significant. To verify this, we conducted a t-test on the test scores, which yielded a t-statistic of -1.706 and a p-value of 0.139. Given that the p-value is larger than the significance level of 0.05, we cannot reject the null hypothesis that there is no significant difference between the performance of UNI and Virchow2. The performance of both models is essentially comparable, suggesting that their pre-trained embeddings can effectively capture relevant features for skin pathology classification.

Logistic regression is the best classifier for both UNI and Virchow2, with accuracies reaching 88% and 90%, respectively. This indicates that the relationship between the extracted features and the classification label is primarily linear. Some ensemble methods, such as Gradient Boosting and Random Forest, achieve the second-highest tier of accuracy. The kNN classifier shows the smallest performance difference (1%) between UNI and Virchow2, indicating that their embedding spaces may have similar local feature structures. Naive Bayes demonstrates the most significant improvement (a 6% increase) when moving from UNI to Virchow2, possibly becauseVirchow2's embeddings better satisfy the independence assumption of the Naive Bayes classifier( i.e., assuming all features are independent of each other).

This observation also aligns with our intuitive understanding of skin diseases, where the correlation between characteristics of different skin disease types is weak. An interesting point to note is that the rankings of classifiers for both models are consistent, suggesting that the choice of classifier remains important regardless of the fundamental model used.

**Table 3**: Performance comparison of classifiers on slide-level embeddings from UNI and Virchow2 foundation models (F1 score in three categories).

| Classifier | Categories | F1 score (UNI) | F1 score (Virchow2) |
|---|---|---|---|
| k-Nearest Neighbor (kNN) | basaloid | 0.72 | 0.76 |
|  | melanocytic | 0.85 | 0.84 |
|  | squamous | 0.86 | 0.88 |
| Decision Tree | basaloid | 0.55 | 0.68 |
|  | melanocytic | 0.79 | 0.77 |
|  | squamous | 0.78 | 0.80 |
| Gradient Boosting | basaloid | 0.77 | 0.86 |
|  | melanocytic | 0.90 | 0.88 |
|  | squamous | 0.89 | 0.91 |
| Random Forest | basaloid | 0.69 | 0.82 |
|  | melanocytic | 0.87 | 0.88 |
|  | squamous | 0.86 | 0.90 |
| Logistic Regression | basaloid | 0.84 | 0.89 |
|  | melanocytic | 0.88 | 0.90 |
|  | squamous | 0.90 | 0.91 |
| Naive Bayes | basaloid | 0.67 | 0.80 |
|  | melanocytic | 0.85 | 0.86 |
|  | squamous | 0.82 | 0.87 |
| AdaBoost | basaloid | 0.69 | 0.81 |
|  | melanocytic | 0.86 | 0.86 |
|  | squamous | 0.83 | 0.87 |

**Table 3** presents the F1 scores of classifiers by category, providing a balanced view of precision and recall. Notably, the squamous cell class consistently achieves the highest F1 score across most classifiers, followed closely by the melanocyte class, while the basaloid class may be attributed to its smaller population size compared to the other two classes.

The advantage of Virchow2 over UNI is particularly evident in basaloid classification, with substantial improvements across all classifiers. For instance, the F1 score for basaloid cell lesions using Naive Bayes increases from 0.67 for UNI to 0.80 with Virchow2, an improvement of 13%. Logistic regression demonstrates strong performance across all classes, achieving the highest F1 scores in the basaloid (0.89) and melanocyte (0.90) classes with Virchow2, as well as the highest score in the squamous cell class (0.91).

Next, we will use the logistic regression classifier to further evaluate the Virchow2 model through the ROC curve and precision-recall curve.

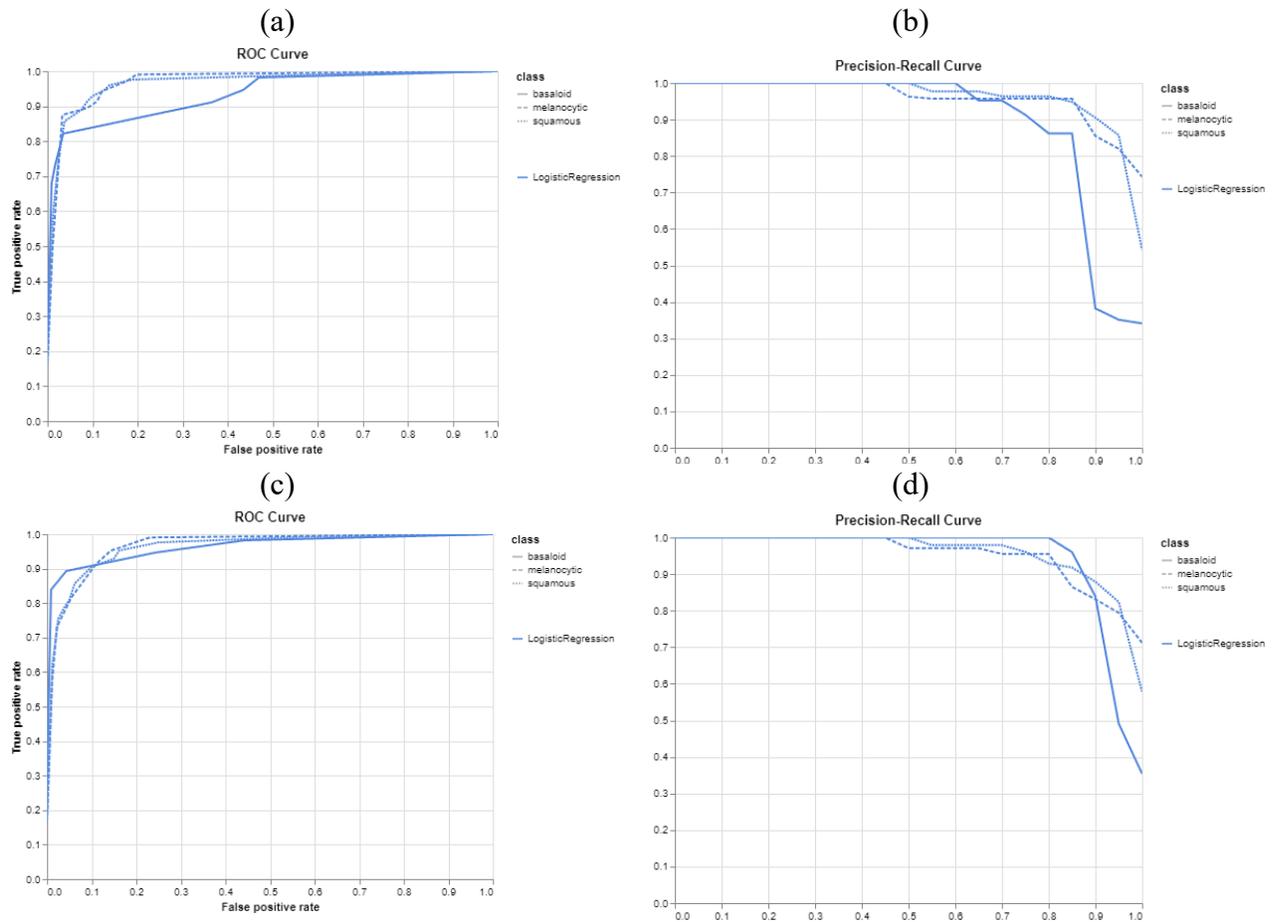

**Figure 7**: Comparative Analysis of UNI and Virchow2 Foundation Models for Multi-class Dermatopathological Classification.
(a) Top-left: ROC Curve for UNI; (b) Top-right: Precision-Recall Curve for UNI; (c) Bottom-left: ROC Curve for Virchow2; (d) Bottom-right: Precision-Recall Curve for Virchow2.

**Figure 7** above shows the comparative analysis of UNI and Virchow2 foundation models for multi-class dermatopathological classification. The ROC curves for both models show excellent classification performance, with the curves tightly hugging the upper left corner. This strong performance is also confirmed by the AUROC values: 96.4% for UNI and 97.0% for Virchow2. While both models demonstrate high accuracy, the 0.6 percentage point improvement achieved by Virchow2 reflects a slight improvement in classification ability.

Notably, the ROC curve for Virchow2 exhibits tighter clustering and is closer to the ideal upper left corner compared to UNI. This indicates that Virchow2 may offer superior and more balanced performance across all three categories (basaloid, melanocytes, and squamous), with the improvement particularly driven by the basaloid class.

The precision-recall curves show that Virchow2 maintains slightly higher precision across a range of recall values for all classes. In contrast, the precision-recall curves for UNI show a more pronounced drop in precision at higher recall levels, particularly for the basaloid class. Virchow2, however, maintains consistently higher precision even at higher recall rates.

Table 4: Statistical Comparison of UNI and Virchow2 Models.

| Category | DeLong's Test | | Venkatraman's Test | |
| --- | --- | --- | --- | --- |
| | Z-Score | P-Value | ROC Difference | P-Value |
| Basaloid | -1.3934 | 0.1635 | 0.0250 | 1.0000 |
| Melanocytic | -0.6225 | 0.5336 | 0.0067 | 1.0000 |
| Squamous | -0.5369 | 0.5913 | 0.0079 | 1.0000 |

Now, we assess the statistical significance of this difference. Two statistical tests were conducted: DeLong's test for comparing AUROCs and Venkatraman's test for comparing ROC curves. The results are shown in **Table 4**. DeLong's test produced p-values of 0.163, 0.534, and 0.591 for the basaloid, melanocytic, and squamous categories, respectively. Similarly, Venkatraman's test yielded p-values of 1.000 for all three categories. Since p-values greater than 0.05 indicate no statistically significant differences, both tests confirm that the differences in ROC curves and AUROCs between UNI and Virchow2 are not statistically significant, despite Virchow2 demonstrating numerical improvements in accuracy over UNI.

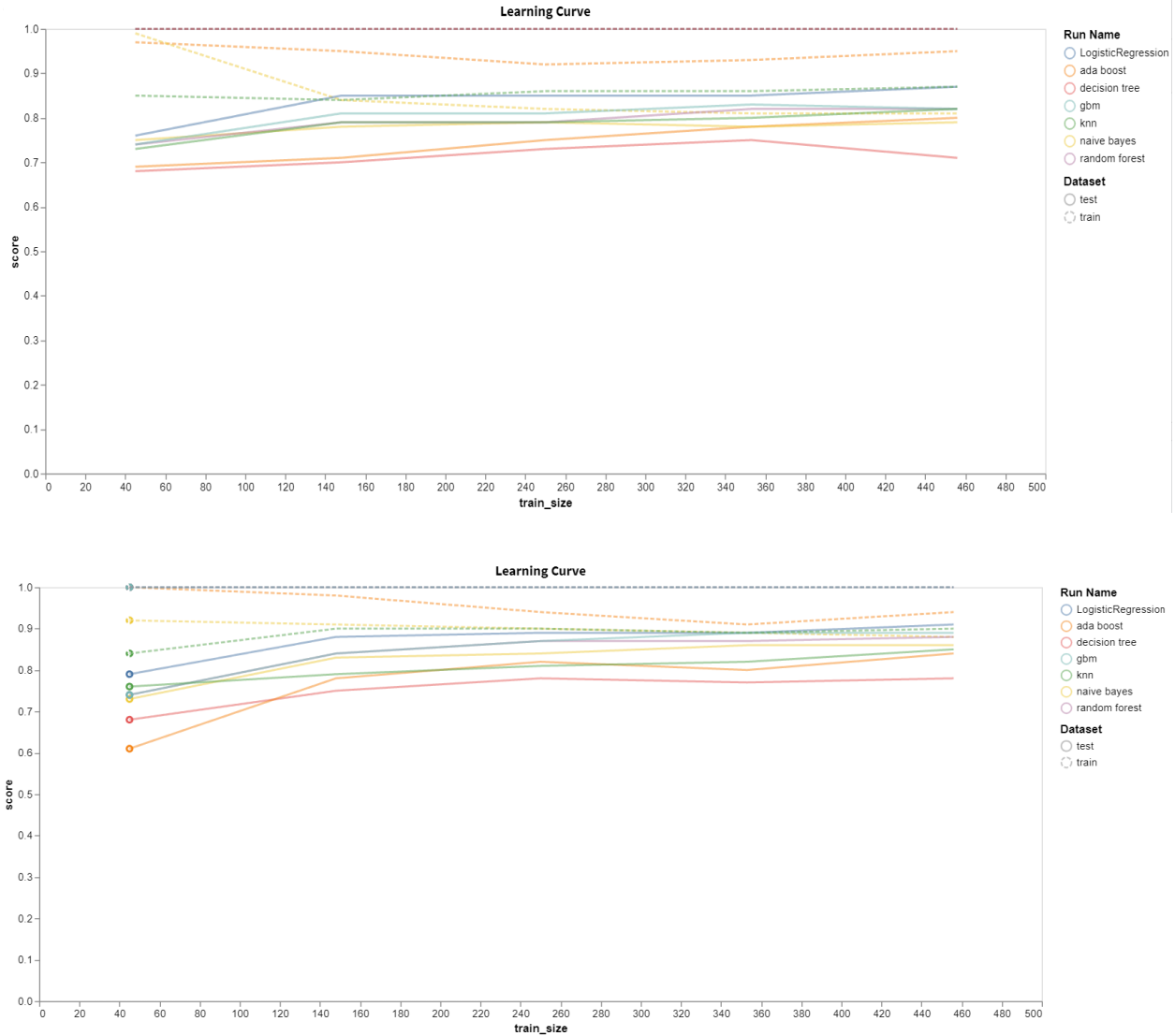

**Figure 8**: Learning Curve of training set size for UNI (above) and Virchow2 (below)

Finally, we analyzed the learning curves with respect to training set size. The learning curves plot the model's accuracy on training and test datasets as a function of training set size, helping to identify potential overfitting or underfitting issues (see **Figure 8**). As the training set t increases, Virchow2's curve demonstrates greater stability and less fluctuation compared to UNI, indicating better generalization ability and lower sensitivity to overfitting.

We also observed that both models reached performance convergence at around 140 training samples, indicating that the models can effectively learn data patterns even with a limited dataset size. Among the classifiers, logistic regression achieves the best performance for both foundation models, suggesting that the classes are linearly separable in the feature space and validating logistic regression as a suitable choice. Importantly, neither model shows significant signs of overfitting, indicating that the model complexities are appropriate for our task.

## DISCUSSION

This study demonstrates the potential for foundation models, specifically UNI and Virchow2, for the automated classification of WSIs into dermatopathological categories. While the rapid digitization of pathological slides offers opportunities for computational analysis with large foundational models, it also presents challenges, including the high resolution of WSIs and the complexity of histopathological features. By leveraging pre-trained models as feature extractors, we demonstrate a promising framework for predicting skin disease types, showcasing the broader applicability of these approaches in medical image analysis.

UNI and Virchow2 were evaluated for their performance in classifying WSIs into three diagnostic categories: melanocytic, basaloid, and squamous lesions. While Virchow2 demonstrated slight numerical improvements over UNI, with accuracies ranging from 77% to 90% compared to UNI's 76% to 88%, statistical analyses (DeLong's and Venkatraman's tests) reveal that these differences were not statistically significant. This suggests that both models are robust in capturing skin histopathological features, offering reliable performance for this application. Logistic regression emerged as the best-performing classifier for both models, achieving 88% accuracy with UNI and 90% with Virchow2, reflecting the predominantly linear relationship between extracted features and classification outcomes. This result underscores the interpretability and clinical relevance of these models in predicting skin disease classifications.

Both models demonstrated efficient learning curves, reaching performance convergence at around 140 training samples. This indicates that the models can learn effectively from limited datasets, which is particularly valuable in medical imaging applications where large annotated datasets are often difficult and time-consuming to obtain. Additionally, Virchow2's learning curves showed greater and lower sensitivity to overfitting, further supporting its generalizability in clinical workflows.

Despite these promising findings, several limitations warrant discussion. First, the mean aggregation strategy employed to combine patch-level features does not capture spatial relationships between patches, which are critical for preserving tissue structure and context. Future studies should explore advanced aggregation methods, such as attention-based mechanisms or graph neural networks, to better account for these spatial relationships.

Second, our dataset was derived from a single institution and included only three diagnostic categories, which may limit the generalizability of our results. Expanding the dataset to encompass additional skin disease categories and incorporating data from multiple institutions is essential for improving the robustness and clinical applicability of these models.

Third, while we acknowledged the potential variability in staining protocols, this study did not extensively address the impact of such variability on model performance. Future research should systematically evaluate how differences in staining and imaging protocols affect model predictions and explore techniques such as stain normalization or domain adaptation to mitigate these challenges. These efforts will be crucial for ensuring consistent and reliable performance across diverse clinical settings.

In summary, our findings underscore the utility of foundation models like UNI and Virchow2 in automating dermatopathology workflows. However, addressing the outlined limitations will be essential for translating these methods into robust, real-world clinical tools.

## CONCLUSION

This study evaluated two foundation models, UNI and Virchow2, for classifying WSIs into dermatopathological categories using machine learning classifiers. By leveraging these pre-trained models as feature extractors, we demonstrated an effective pipeline for generating slide-level predictions. Both models demonstrated strong performance, with Virchow2 achieving the highest accuracy (90%) using logistic regression, though the difference was not statistically significant. Our findings provide a scalable framework for automating diagnostic workflows in dermatopathology and lay the groundwork for future advancements in the field.

## ACKNOWLEDGEMENTS

We would like to express our gratitude to Dennis Murphree, PhD, from Mayo Clinic's Dermatology Department, for his invaluable mentorship and guidance throughout this project. We are also deeply appreciative of Professor Santiago Romero-Brufau and Teaching Fellow Tony Chen from the HDSC 325 course at Harvard T.H. Chan School of Public Health for their expert guidance and insightful feedback. Additionally, we thank the Mayo Clinic Dermatology Department team for providing access to the remote server and for their assistance with the segmentation and labeling of the patches used in our study. This capstone project is part of our coursework at the Harvard T.H. Chan School of Public Health, and we are grateful for the collaboration and resources that made this work possible.

## CITATIONS